\documentclass{sjtudenglab}
\usepackage{amsmath}
\usepackage{enumerate} 
\usepackage{algorithm}
\usepackage{algpseudocode}
\usepackage{amsfonts}
\usepackage{amsthm}
\usepackage{newtxtt}
\usepackage{diagbox} 
\usepackage{colortbl}
\usepackage{amssymb}
\usepackage{xspace}
\usepackage{wrapfig}
\usepackage{adjustbox}
\usepackage{tabularx}
\usepackage{booktabs}
\usepackage{mathtools}
\usepackage{tikz}
\usepackage{enumitem}
\usepackage{silence}
\usepackage{dsfont}
\usepackage[table]{xcolor}
\usepackage[dvipsnames]{xcolor}
\usepackage{multirow}
\usepackage{makecell}
\usepackage{xfakebold}
\usepackage{natbib}
\usepackage{siunitx}
\usepackage{cleveref}
\usepackage{thmtools}
\usepackage{float}

\usepackage{amsmath,amsfonts,bm}

\def\eqref#1{equation~\ref{#1}}

\def\1{\bm{1}}

\DeclareMathAlphabet{\mathsfit}{\encodingdefault}{\sfdefault}{m}{sl}
\SetMathAlphabet{\mathsfit}{bold}{\encodingdefault}{\sfdefault}{bx}{n}

\usepackage{parskip}

\usepackage{natbib}
\usepackage{amsfonts,bm}

\usepackage{amsmath, amssymb,mathrsfs, amsthm, mathtools}
\usepackage{xspace}
\usepackage{enumitem}
\usepackage{lipsum}
\usepackage{xpatch}
\usepackage{mathtools}
\usepackage{dsfont}
\usepackage{pgf,tikz}
\usetikzlibrary{positioning,matrix,patterns}
\usepackage{caption}
\usepackage{graphicx}
\usepackage{makecell}
\usepackage{multirow}
\usepackage[mathscr]{euscript}
\definecolor{hanblue}{rgb}{0.27, 0.42, 0.81}
\definecolor{deepred}{HTML}{900C3F}
\definecolor{deepgreen}{HTML}{2F6960}
\hypersetup{
    colorlinks=true,
    linkcolor=hanblue,
    urlcolor=hanblue,
    citecolor=hanblue,
    anchorcolor=hanblue}

\declaretheoremstyle[
  headfont=\sffamily\bfseries,
]{sansserif}

\theoremstyle{sansserif}

\theoremstyle{definition}

\theoremstyle{sansserif}

\theoremstyle{remark}

\DeclarePairedDelimiter\abs{\lvert}{\rvert}%
\DeclarePairedDelimiter\norm{\lVert}{\rVert}%

\makeatletter
\let\oldabs\abs
\def\abs{\@ifstar{\oldabs}{\oldabs*}}
\let\oldnorm\norm
\def\norm{\@ifstar{\oldnorm}{\oldnorm*}}
\makeatother

\definecolor{textgray}{HTML}{6E6E73}
\usetikzlibrary{positioning, calc}
\usetikzlibrary{decorations.pathmorphing}

\makeatletter
\patchcmd{\wrong@fontshape}{\@gobbletwo}{}{}{}
\makeatother
\numberwithin{equation}{section} 
\tcbuselibrary{minted}
\setminted[python]{
  linenos,
  breaklines,
  fontsize=\footnotesize,
  xleftmargin=2em
}

\makeatletter
\AtBeginDocument{
  \urlstyle{sf}
  
}
\makeatother

\floatstyle{plain}
\newfloat{algorithmgroup}{tbp}{loag}
\floatname{algorithmgroup}{Algorithm Group}

\def\abbr{CineMobile\xspace}

\renewcommand{\eqref}[1]{\textup{(\ref{#1})}}

\definecolor{light}{RGB}{125, 125, 125}
\crefname{tcb@cnt@pbox}{code}{code}
\Crefname{tcb@cnt@pbox}{Code}{Code}
\crefname{assumption}{assumption}{assumption}
\Crefname{assumption}{Assumption}{Assumptions}

\newtcolorbox[auto counter]{pbox}[2][]{
  colback=white,
  title=Code~\thetcbcounter: #2,
  #1,fonttitle=\sffamily,
  fontupper=\sffamily,
  arc=2pt,
  colframe=bgcolor,
  coltitle=fgcolor,
  colbacktitle=bgcolor,
  toptitle=0.25cm,
  bottomtitle=0.125cm
}

\makeatletter
\newcommand\applefootnote[1]{%
  \begingroup
  \renewcommand\thefootnote{}%
  \renewcommand\@makefntext[1]{\noindent##1}%
  \footnote{#1}%
  \addtocounter{footnote}{-1}%
  \endgroup
}
\makeatother

\definecolor{cverbbg}{gray}{0.90}

\title{CineMobile: On-Device Image-to-Video Diffusion for Cinematic Camera Motion Generation}

\author{
Xuyao Huang$^{1,*}$,
Zelai Deng$^{2,*}$,
Xu Wang$^{1}$,
Xizhong Xiao$^{3}$,
Zhijie Deng$^{1}$
}

\affiliation{\small
  $^{1}$Shanghai Jiao Tong University,
  $^{2}$Nankai University,
  $^{3}$Transsion}

\abstract{
The growing demand for image-to-video creation on mobile devices has increasingly focused on cinematic motion effects like \emph{bullet time}, \emph{dolly zoom}, \emph{slow motion}, etc. While Diffusion Transformers (DiTs) exhibit strong performance in video generation, their large parameter sizes and multi-step iterative denoising processes lead to substantial computational overhead, making efficient generation on mobile devices challenging. We propose \textbf{CineMobile} to bridge the gap. 
In particular, CineMobile adopts a three-fold optimization strategy: (1) leveraging a distillation-guided pruning approach to derive a compact yet efficient model that retains the essential video generation capabilities required for cinematic effects; (2) optimizing the compressed model into a 4-step generator via a combination of diffusion distillation and reinforcement learning; (3) employing a hybrid post-training quantization strategy to compress the model footprint to under 1 GB. 
Experimental results show that compared to the teacher model with the Wan 2.1 architecture, CineMobile achieves a \textbf{40$\times$ speedup} in generation while maintaining comparable visual quality. 
Specifically, \abbr generates 49-frame 480p videos with a per-step denoising latency of \textbf{0.6s} on an NVIDIA H200 GPU and \textbf{20s} on the MediaTek Dimensity 8400 Ultimate 5G platform, with a peak memory usage of \textbf{1.8 GB}, demonstrating its practical applicability for mobile-based image-to-video creation.

}

\metadata[Correspondence]{\sffamily Zhijie Deng: \url{zhijied@sjtu.edu.cn}} 
\date{\sffamily\today}

\AtBeginDocument{ 
  \fancyheadoffset[R]{-0.3cm}
  \fancyhead[R]{\raisebox{9pt}{\includegraphics[height=35pt]{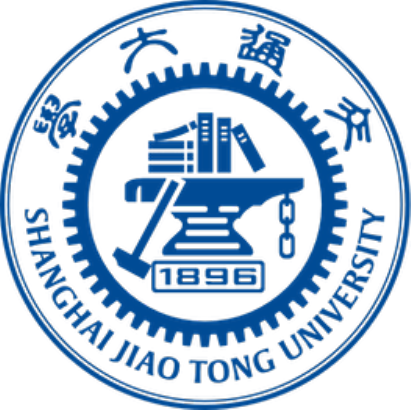}}}
}

\begin{document}

\maketitle

\begin{figure}[ht]
    \centering
    \includegraphics[width=\linewidth]{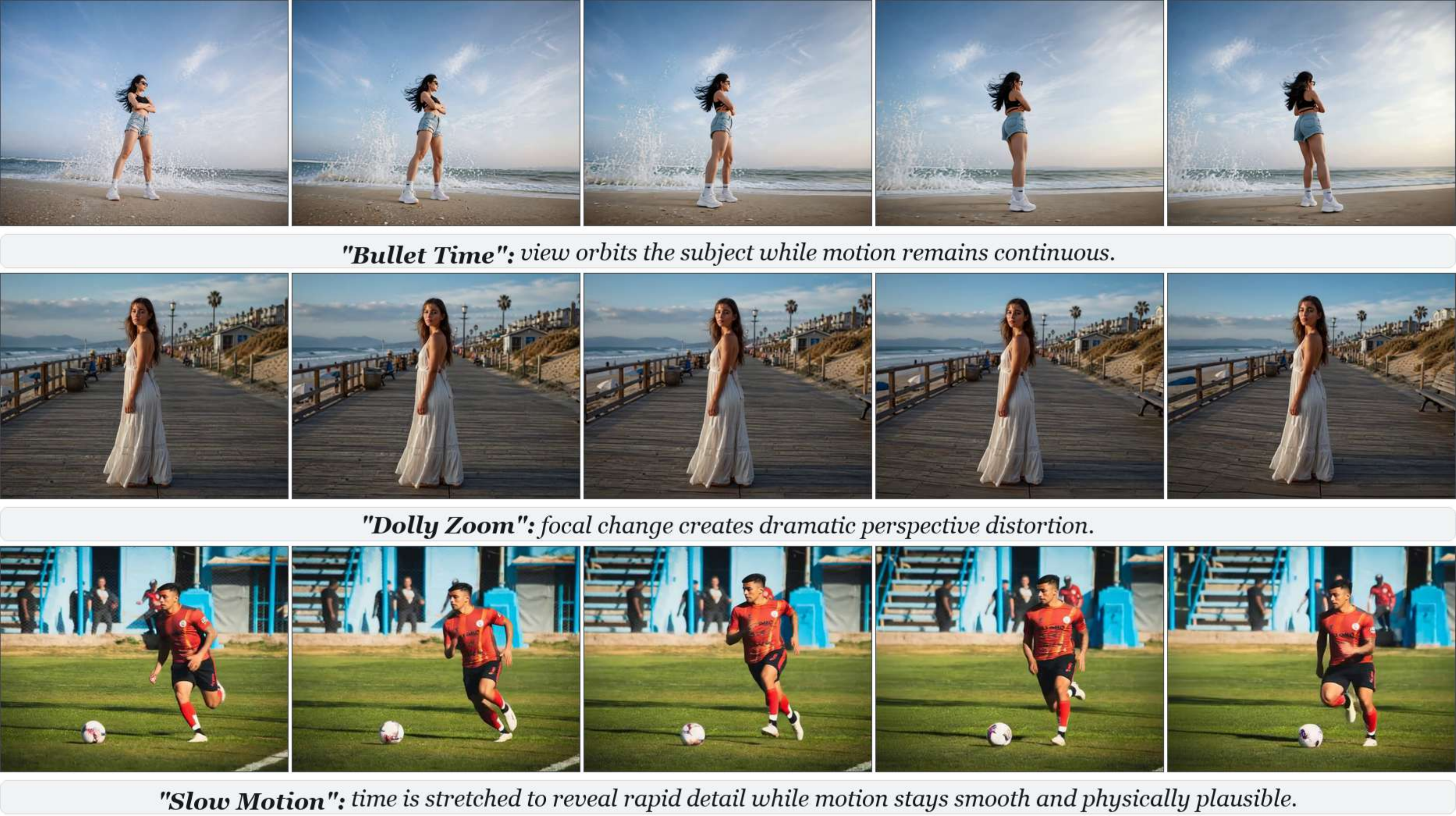}
    \caption{Bullet time, dolly zoom, and slow motion videos generated by CineMobile. CineMobile can produce continuous cinematic camera motion while preserving subject identity and scene consistency.}
    \label{fig:abs}
\end{figure}

\section{Introduction}
\label{sec:introduction}

The widespread adoption of smartphones for visual content creation is driving an increasing demand for mobile image-to-video (I2V) generation. Meanwhile, recent video generation models—such as Kling 2.5 Turbo~\citep{klingteam2025klingomnitechnicalreport}, Wan2.7~\citep{wan2025wanopenadvancedlargescale}, Vidu Q3~\citep{bao2024viduhighlyconsistentdynamic}, and Seedance 2.0~\citep{seedance2026seedance20advancingvideo}—have demonstrated remarkable capabilities in generating high-quality, realistic videos, with Diffusion Transformers (DiTs) emerging as the leading architecture underpinning these advances.

Despite their strong performance, DiTs suffer from huge parameter sizes and slow generation speed, limiting their practicality for cost-effective video creation~\citep{Kahatapitiya_2025_ICCV}.
For example, Wan2.1-I2V-14B~\citep{wan2025wanopenadvancedlargescale} requires 240 seconds to generate a 49-frame video at 832 × 480 resolution with 40 inference steps on an NVIDIA H200 GPU, making mobile deployment infeasible. 
Furthermore, existing acceleration methods for deploying video generation models on mobile devices mainly focus on U-Net architectures~\citep{10.1007/978-3-319-24574-4_28}, leaving DiT models largely underexplored~\citep{zhang2025mobilei2vfasthighresolutionimagetovideo,wu2025tamingdiffusiontransformerefficient}. Considering that I2V models struggle to faithfully preserve prompt semantics under aggressive acceleration~\citep{Lv_2025_ICCV}, we focus on camera motion effects with more structured and controllable motion targets.

This paper proposes \textbf{\abbr} to enable DiT-based cinematic camera motion generation on mobile devices, based on a combination of structured pruning, step distillation, and hybrid precision quantization. 
Concretely, we first extend the Pluggable Pruning with Contiguous Layer Distillation (PPCL)~\citep{ma2026pluggablepruningcontiguouslayer} method, originally designed for text-to-image models, to prune image-to-video DiTs. 
Considering our prioritization of preserving fine portrait details and stabilizing camera control, 
we identify that it is better to perform structured depth pruning while maintaining the hidden dimension of the backbone unchanged. 

\begin{figure}[t]
    \centering
    \includegraphics[width=\linewidth]{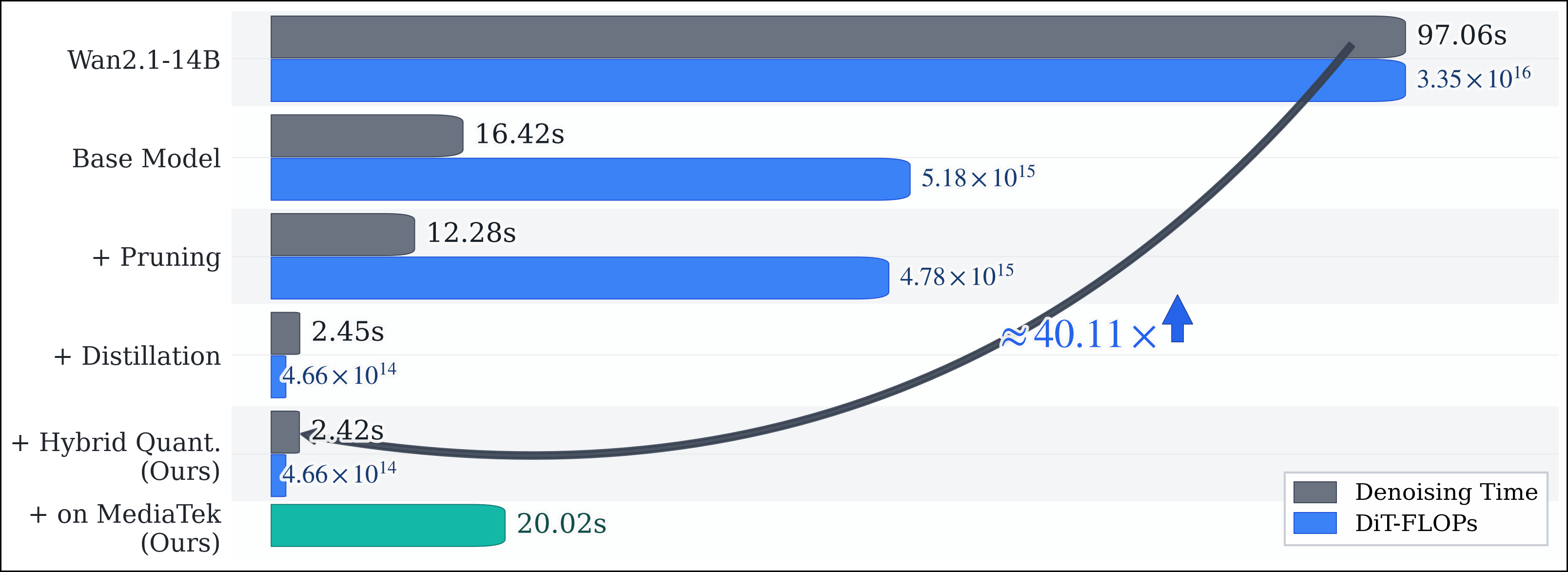}
    \caption{Denoising time and DiT-FLOPs comparison of CineMobile across different acceleration stages. Wan2.1-14B is used as the teacher model, requiring 97.06s denoising time and $3.35 \times 10^{16}$ DiT-FLOPs. Starting from the base model Wan2.1-v1.1-Fun-1.5B, pruning reduces the model complexity and brings a moderate speedup, reducing the denoising time from 16.42s to 12.28s. Distillation further enables 4-step generation, leading to the main efficiency improvement with 2.45s denoising time. After applying hybrid precision weight quantization, CineMobile achieves \textbf{40.11×} faster DiT denoising and \textbf{71.89×} lower DiT-FLOPs than the teacher model. On the MediaTek Dimensity 8400 Ultimate 5G platform, CineMobile achieves a per-step denoising latency of 20.02s.
    }
    \label{fig:speed}
\end{figure}

After pruning, we introduce a supervised fine-tuning warm-up stage to restore the student's fundamental image-to-video generation capability and adapt it to the target camera-motion distribution. 
For step distillation, we adapt AdvDMD~\citep{wang2026advdmdadversarialrewardmeets}, which integrates reinforcement learning (RL)~\citep{li2018deepreinforcementlearningoverview} into distribution matching distillation (DMD)~\citep{Yin_2024_CVPR} for text-to-image generation, to I2V DiTs. 

The involved RL process enables \abbr to generate high-quality videos with a small number of denoising steps.
To reduce the on-device memory cost, we adopt a hybrid precision quantization strategy. 
By quantizing FFN weights to 4-bit and the remaining components to 8-bit~\citep{lightx2v}, \abbr reduces the model memory footprint to under 1 GB while preserving visual quality.

To assess the capability of \abbr in generating cinematic camera and temporal effects, we evaluate it on three representative cinematic motion effects: bullet time, dolly zoom, and slow motion. These effects cover distinct forms of cinematic motion control with clear motion patterns, making them suitable for studying controllable and efficient on-device I2V generation.
We evaluate the generated videos on high-quality portrait and motion data using the VBench~\citep{Huang_2024_CVPR} evaluation protocol and human evaluation. 
As shown in \Cref{fig:speed}, with only 4 denoising steps and a model footprint of less than 1 GB, \abbr achieves a $40\times$ speedup over the teacher model in denoising time, and a per-step latency of 20s on the MediaTek Dimensity 8400 Ultimate 5G platform.
In \Cref{tab:vbench}, following the VBench evaluation protocol, \abbr maintains generation quality comparable to the teacher model across all three cinematic motion-effect scenarios, with total scores of 88.35, 89.30, and 88.05 on bullet time, dolly zoom, and slow motion, respectively, closely matching the teacher model's 89.27, 89.96, and 88.51.

\begin{figure}
    \centering
    \includegraphics[width=\linewidth]{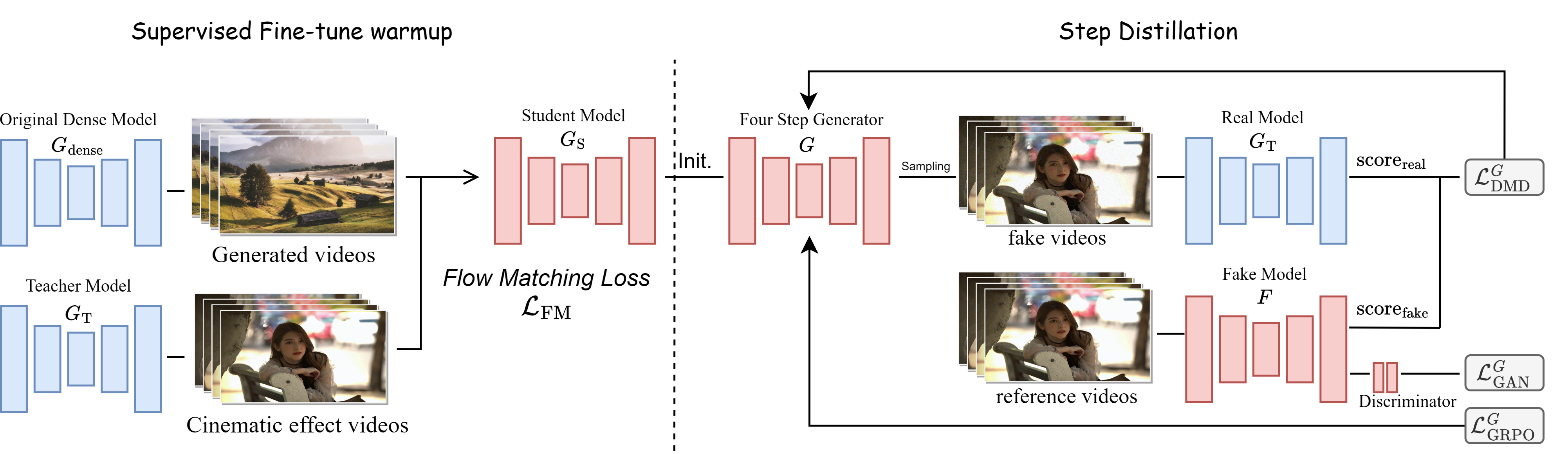}
    \caption{Overview of the step distillation pipeline.
    In the supervised fine-tuning warm-up, the student model is trained with the flow-matching objective on videos generated by the original dense model and cinematic-effect videos produced by the teacher model.
    The warm-up checkpoint then initializes a 4-step generator, which is further refined through adversarial distillation.
    In this stage, real model \(G_T\) and an online fake model \(F\) provide distribution-matching signals, while a lightweight discriminator supplies adversarial supervision and rewards for GRPO.}
    \label{fig:distill}
\end{figure}

\section{Related Work}
\label{gen_inst}

\paragraph{Video Diffusion Models.}
Video generation has rapidly evolved from early video diffusion and latent video models~\citep{NEURIPS2022_39235c56,he2023latentvideodiffusionmodels,zhou2023magicvideoefficientvideogeneration} to stronger large-scale foundations such as Stable Video Diffusion~\citep{blattmann2023stablevideodiffusionscaling}, Lumiere~\citep{10.1145/3680528.3687614}, CogVideoX~\citep{yang2025cogvideoxtexttovideodiffusionmodels}, Open-Sora~\citep{zheng2024opensorademocratizingefficientvideo}, LTX-Video~\citep{hacohen2024ltxvideorealtimevideolatent}, and Wan~\citep{wan2025wanopenadvancedlargescale}. These models improve visual fidelity, motion coherence, and scalability through advances in temporal modeling, latent compression, and backbone design~\citep{wang2026surveyvideodiffusionmodels}. From the perspective of task formulation, existing video diffusion models can be broadly organized into text-to-video generation~\citep{10.1007/978-3-031-72986-7_23,hong2023cogvideo,ICLR2025_3ab228c4,NEURIPS2024_8a57aa8e,hassan2026anchoredvideogenerationdecoupling,zhang2025riset2vrephrasinginjectingsemantics}, image-to-video generation~\citep{blattmann2023stablevideodiffusionscaling,zhou2024storydiffusionconsistentselfattentionlongrange,10.1145/3641519.3657497,ren2024consistiv,ICLR2025_5fba7090}, and specific motion-controlled generation~\citep{ren2024customize,wu2024motionboothmotionawarecustomizedtexttovideo,zeng2023makepixelsdancehighdynamic,zhao2023motiondirectormotioncustomizationtexttovideo,Geng_2025_CVPR,wang2025atitrajectoryinstructioncontrollable}. Text-to-video models emphasize semantic alignment and open-ended video synthesis at scale~\citep{singer2022makeavideotexttovideogenerationtextvideo,li2023videogenreferenceguidedlatentdiffusion}, while image-to-video models focus more on preserving subject identity, scene layout, and appearance consistency under temporal evolution~\citep{10.1145/3641519.3657407,10.1145/3641519.3657497}.
Despite this progress, most high-quality video diffusion models still depend on large backbones, server-class hardware, and multi-step sampling, limiting their practical deployment on edge devices~\citep{zheng2025diffusionmodelsedgechallenges}. This limitation becomes more critical in image-to-video settings with structured camera effects, where controllability and efficiency must be achieved simultaneously~\citep{zheng2025diffusionmodelsedgechallenges,shao2026efficientvideodiffusionmodels,wu2025tamingdiffusiontransformerefficient}.

\begin{figure}[!t]
    \centering
    \includegraphics[width=\linewidth]{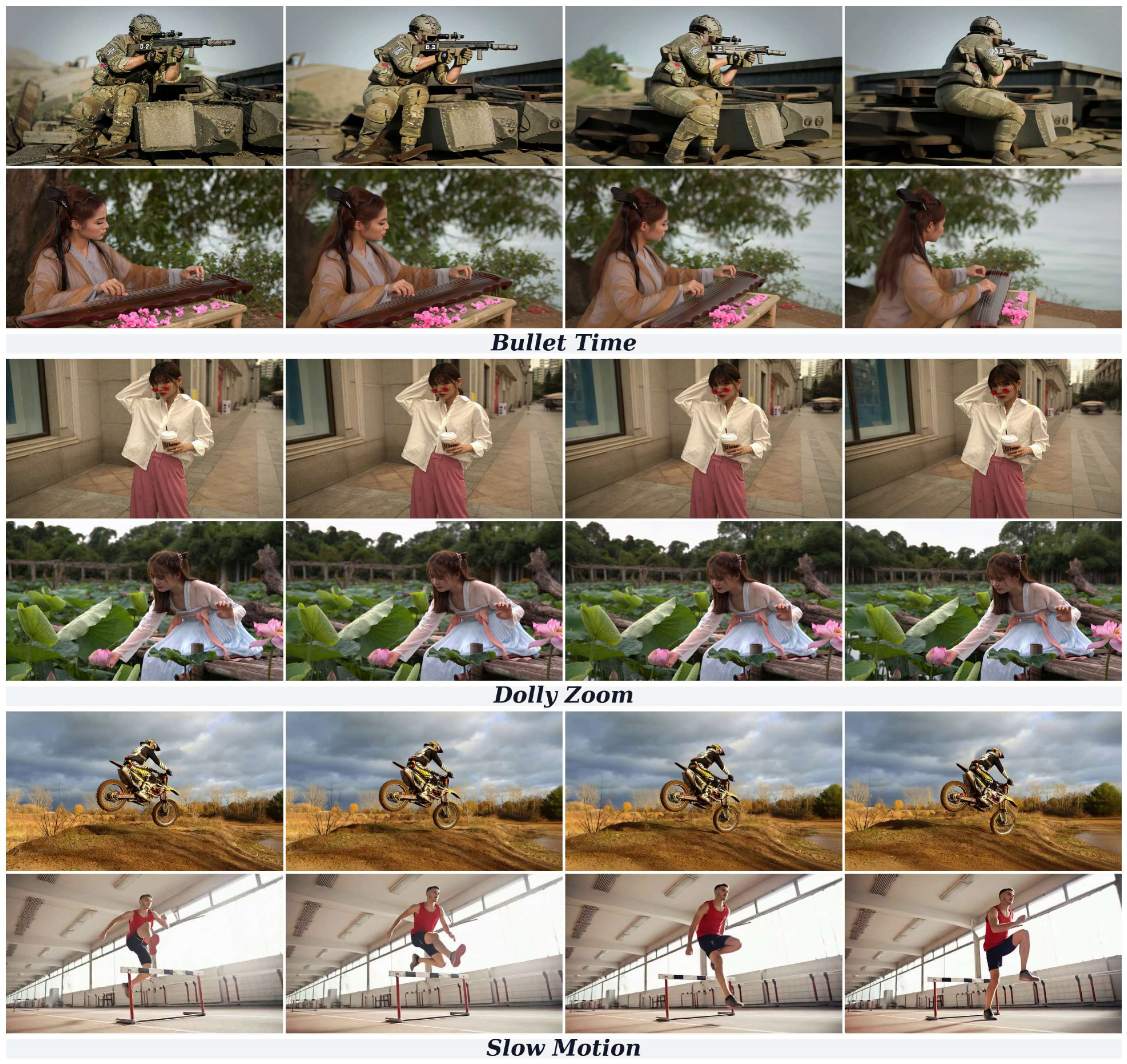}
    \caption{Qualitative examples of CineMobile across three cinematic camera effects. }
    \label{fig:effects}
\end{figure}

\paragraph{On-Device Models.}
Recent work has started to bring video generation to mobile devices~\citep{kim2025ondevicesoraenablingtrainingfree,wu2025snapgenvgeneratingfivesecondvideo,zhang2025mobilei2vfasthighresolutionimagetovideo,yahia2024mobilevideodiffusion,wu2025tamingdiffusiontransformerefficient}. Mobile Video Diffusion compresses image-to-video generation through resolution reduction, temporal multi-scaling, pruning, and adversarial fine-tuning~\citep{yahia2024mobilevideodiffusion}. On-device Sora improves mobile deployment with denoising reduction, temporal token merging, and dynamic loading~\citep{kim2025ondevicesoraenablingtrainingfree}. SnapGen-V shows that interactive mobile generation can be achieved by combining an efficient spatial backbone, mobile-oriented temporal design, and adversarial low-step training~\citep{wu2025snapgenvgeneratingfivesecondvideo}. Taming Diffusion Transformer for Efficient Mobile Video Generation in Seconds further extends this line to DiT-based models with highly compressed VAE design, sensitivity-aware pruning, and mobile-oriented step distillation~\citep{wu2025tamingdiffusiontransformerefficient}. These studies indicate that practical on-device video generation depends on joint optimization of architecture, compression, memory, and sampling.

\paragraph{Step Distillation.} Reducing denoising steps is a central direction for efficient diffusion generation~\citep{Du_2025_ICCV,salimans2022progressive}. Early acceleration methods include DDIM~\citep{song2022denoisingdiffusionimplicitmodels} and Progressive Distillation~\citep{salimans2022progressive}, followed by stronger few-step paradigms such as Latent Consistency Models~\citep{luo2023latentconsistencymodelssynthesizing}, Adversarial Diffusion Distillation~\citep{10.1007/978-3-031-73016-0_6}, DMD~\citep{Yin_2024_CVPR}, and DMD2~\citep{yin2024improveddistributionmatchingdistillation}. In video generation, VideoLCM~\citep{wang2023videolcmvideolatentconsistency}, AnimateLCM~\citep{wang2024animatelcmcomputationefficientpersonalizedstyle}, T2V-Turbo~\citep{NEURIPS2024_8a57aa8e}, and T2V-Turbo-v2~\citep{ICLR2025_e68af7d8} adapt these ideas to the temporal setting and achieve competitive quality with only a few sampling steps. Recent mobile systems combine step distillation with hardware-aware model design: Mobile Video Diffusion approaches single-step inference through adversarial fine-tuning~\citep{yahia2024mobilevideodiffusion,kim2025ondevicesoraenablingtrainingfree}, SnapGen-V reduces generation to 4 steps~\citep{wu2025snapgenvgeneratingfivesecondvideo}, and Taming Diffusion Transformer develops adversarial step distillation for compressed mobile DiTs~\citep{wu2025tamingdiffusiontransformerefficient}. These works are especially relevant to our setting, where efficient on-device image-to-video generation requires few-step sampling together with aggressive model compression.

\section{Method}
\label{sec:method}
Our objective is to build an I2V model that can efficiently generate different camera effect videos on mobile devices. 
To support lightweight deployment and diverse effects, \abbr uses a shared backbone with effect-specific LoRA modules.
To reduce the memory footprint, we first prune the shared DiT backbone in \Cref{method:pruning}. 
After pruning, we restore the generative capability through a two-stage fine-tuning procedure, and then distill the model into a 4-step generator in \Cref{method:distillation}.
Finally, we obtain the final model through hybrid post-training quantization in \Cref{method:quantization}.

\subsection{Structured Depth Pruning}
\label{method:pruning}

Directly pruning video DiTs can easily disrupt temporal modeling, leading to identity drift and severe degradation of fine-grained details, especially for cinematic motion generation that requires strict temporal consistency and camera trajectory smoothness. Given the strong performance of PPCL on text-to-image DiTs, we adapt it to video DiTs with specific designs for video temporal modeling and portrait detail preservation. PPCL~\citep{ma2026pluggablepruningcontiguouslayer} first detects redundant contiguous layer intervals in DiTs, and then compresses the model through depth-wise and width-wise pruning with distillation.
Empirically, we find that width pruning can disturb the pretrained temporal representations of video DiTs by changing their internal feature dimensionality, as detailed in Appendix~\Cref{appendix:pruning}. We adopt a structured depth pruning strategy.
Specifically, our pruning procedure consists of two stages: redundant interval detection and interval distillation.


\paragraph{Redundant interval detection.}
We first train a residual linear probe for each transformer block to estimate whether its transformation can be approximated by a lightweight substitute. 
For a teacher video DiT with blocks $T=\{T_1,T_2,\ldots,T_M\}$, we denote the activation after block $T_i$ as $h_i^T$.
We use residual linear probes as lightweight diagnostics and CKA similarity~\citep{pmlr-v97-kornblith19a} to rank candidate intervals by representation similarity and removable depth.

To adapt interval detection to video generation, we evaluate candidate spans on latent video sequences. We use more than 5,000 calibration samples to obtain stable span estimates that reflect redundancy under temporal motion and spatial appearance interactions.

\paragraph{Interval distillation.}
For each selected span $[u,v]\in\mathcal{P}$, we replace the whole span with one trainable surrogate transformer block $D_u$.
It maintains the same block structure and hidden dimensionality as the base model blocks, and is initialized from the middle block $T_{\lfloor(u+v)/2\rfloor}$ for stable training.

During distillation, the base model is frozen, and only the surrogate blocks are updated.
We use portrait and motion data synthesized by the frozen base model~\citep{wan2025wanopenadvancedlargescale}.
These samples cover facial identity and motion patterns, helping the surrogate blocks retain identity consistency and motion controllability during interval pruning.
For each span, the surrogate block takes the teacher's hidden state at the span input and is trained to reproduce the teacher's hidden state at the span output through normalized hidden-state alignment.
After training, we keep $D_u$ at position $u$ and remove blocks $T_{u+1},\ldots,T_v$.

\subsection{Step Distillation}
\label{method:distillation}

We adopt a two-stage fine-tuning strategy for the pruned model. 
Directly applying step distillation to the pruned model can lead to unstable training and even generation collapse. 
Therefore, we first perform supervised fine-tuning as a warm-up stage, which restores the model's ability to reliably generate motion-effect videos with multi-step sampling. For step distillation, we build upon AdvDMD~\citep{wang2026advdmdadversarialrewardmeets} to improve the quality and stability of few-step generation. 
We extend AdvDMD to video DiTs and use it to distill our model as a 4-step I2V generator.

\paragraph{Supervised fine-tuning warm-up.}
We begin with a supervised fine-tuning warm-up to provide the compressed student with a stable initialization before adversarial optimization. This warm-up consists of two steps under the standard flow-matching objective~\citep{10.1007/978-3-031-70381-2_21}. 
We start by fine-tuning the student model on videos generated by the original dense model to restore its fundamental video generation capability. 
We then use cinematic-effect videos generated by the teacher model to adapt the student model to the distributions of the target motion effects. 
Both steps share the same flow-matching training objective.

\paragraph{Adversarial distillation.}

After the supervised warm-up, we further refine the student through an adversarial distillation stage. 
The student is initialized from the warm-up checkpoint and updated with LoRA adapters. 
As illustrated in \Cref{fig:distill}, we use a frozen real score estimator \(G_T\) and an online fake score estimator \(F\). 
\(G_T\) is initialized from the teacher model with motion-effect LoRA, and provides the teacher-induced score direction. 
\(F\) shares the student's base backbone, with randomly initialized LoRA adapters updated online to track the evolving student distribution.
To provide adversarial realism feedback, we attach a lightweight discriminator head to the fake model \(F\). 
Given an intermediate video latent and its associated I2V condition, \(F\) extracts multi-layer spatio-temporal DiT features, which are mapped by the discriminator heads to realism logits.

For each rollout batch, the 4-step generator \(G\) produces fake videos conditioned on the input frame. Reference videos are generated by the teacher model under the same conditions and serve as motion-effect targets. The frozen real model \(G_T\) and the online fake model \(F\) estimate the real and fake score directions on generated samples, and their discrepancy forms the DMD loss \(\mathcal{L}^{G}_{\mathrm{DMD}}\) for updating the generator. Meanwhile, the discriminator attached to \(F\) provides an adversarial loss \(\mathcal{L}^{G}_{\mathrm{GAN}}\), which encourages generated videos to be judged as reference-like.
The discriminator scores on intermediate latents are further used as GRPO rewards, and \(\mathcal{L}^{G}_{\mathrm{GRPO}}\) is optimized to refine the 4-step trajectory.

We adopt an alternating training schedule: the generator is updated once every five steps, while the fake model and discriminator are updated in the remaining steps to maintain stable score estimation and realism feedback.

\subsection{Hybrid Post-training Quantization}
\label{method:quantization}

After obtaining a generator capable of producing videos in 4 denoising steps, we apply a hybrid post-training quantization strategy to further reduce the runtime memory footprint and latency for on-device generation~\citep{lightx2v}.
Instead of assigning a uniform precision to all layers, we adopt a layer-wise hybrid quantization scheme for the DiT backbone.
We keep activations in $16$-bit precision for numerical stability and use FP8 weights for most quantized linear layers, including attention projections, text-embedding projections, and time-embedding projections.
Since feed-forward networks contain a large fraction of the DiT parameters, the two linear projections in each block use $4$-bit weights with $16$-bit activations.

\section{Experiments}
\label{sec:exp}

In this section, we systematically evaluate the efficiency and performance of \abbr on three cinematic motion effects.

\subsection{Setup}

\paragraph{Training Details.}
For the initial pruning stage, we use more than 5,000 samples to estimate layer redundancy and identify redundant block intervals. 
Based on this analysis, we prune the original DiT backbone to obtain a 27-layer student model, followed by 8k training steps to restore its generation capability.
For the subsequent supervised fine-tuning warm-up, we follow the training implementation of DiffSynth-Studio~\citep{diffsynthstudio2026} and train the student model for 5k steps. 
We use a learning rate of $1\times10^{-4}$ and set the LoRA rank to 128.
During distillation, we use over 10K training samples to optimize the student model, which follows a 4-step denoising trajectory. The student is trained with a learning rate of $1\times10^{-4}$.
We set $\lambda_{\mathrm{adv}}=0.5$ and $\lambda_{\mathrm{D}}=0.005$, and train the discriminator head and fake model with a learning rate of $1\times10^{-5}$.
All experiments are conducted on 8 NVIDIA H200 GPUs.

\paragraph{Model \& Datasets}
To obtain reliable supervision for cinematic motion effect generation, we build teacher models by combining a shared Wan2.1-I2V-14B image-to-video backbone with effect-specific LoRA adapters.
For bullet time and dolly zoom, we use publicly available LoRA adapters tailored to these effects.
For slow motion, we construct the adapter by fine-tuning on slow motion videos derived from motion-intensive base-model outputs through $2\times$ frame interpolation~\citep{reda2022film}.
For the student model, we adopt a Wan2.1-v1.1-Fun-I2V model~\citep{wan2025wanopenadvancedlargescale} with 1.5B parameters as the base model.
For data collection, we use the high-quality portrait dataset PPR10K~\citep{jie2021PPR10K} as the image source for bullet time and dolly zoom. For slow motion, we collect motion-oriented images from public platforms such as Pexels and Pixabay. Each effect-specific training set contains over 10K samples.

\begin{table*}[t]
\centering
\normalsize
\setlength{\tabcolsep}{2.6pt}
\renewcommand{\arraystretch}{1.12}
\caption{Systematic comparison of video generation models on VBench with different parameter sizes and inference steps, including representative publicly available models such as HunyuanVideo-1.5~\citep{wu2025hunyuanvideo15technicalreport}, LTX2.3~\citep{hacohen2026ltx2efficientjointaudiovisual}, and Open-Sora2.0~\citep{zheng2026opensora20trainingcommerciallevel}. 
All models are evaluated on the same test set of 200 portrait images with five videos generated for each image, following the official VBench protocol to compare video generation quality. 
The General panel reports representative publicly available video generation models. 
Wan2.1 is used as the teacher model and Wan2.1-v1.1-Fun as the base model; Ours-Prune, Ours-SFT, and Ours correspond to the pruned-and-finetuned model, the second-stage SFT model, and the final distilled and hybrid-PTQ accelerated model, respectively. 
With only 1.2B parameters and 4 denoising steps, Ours achieves VBench scores close to the teacher model, demonstrating comparable generation quality with substantially higher efficiency. }
\label{tab:vbench}
\resizebox{\textwidth}{!}{
\begin{tabular}{lcccccccccc}
\toprule
\textbf{Model} 
& \textbf{Steps}
& \makecell{\textbf{Params}\\\textbf{(B)}}
& \makecell{\textbf{Subject}\\\textbf{Consis.}} 
& \makecell{\textbf{Background}\\\textbf{Consis.}} 
& \makecell{\textbf{Motion}\\\textbf{Smooth.}} 
& \makecell{\textbf{Aesthetic}\\\textbf{Quality}} 
& \makecell{\textbf{Imaging}\\\textbf{Quality}} 
& \makecell{\textbf{I2V}\\\textbf{Subject}} 
& \makecell{\textbf{I2V}\\\textbf{Background}} 
& \makecell{\textbf{Total}\\\textbf{Score}} \\
\midrule

\multicolumn{11}{c}{\textbf{General}} \\
\midrule
HunyuanVideo-1.5 & 12 & 8.3 & 98.46 & 96.49 & 99.61 & 64.57 & 70.78 & 99.39 & 99.49 & 89.83 \\
LTX2.3           & 12 & 22  & 98.39 & 96.08 & 99.38 & 63.06 & 71.54 & 98.87 & 98.60 & 89.42 \\
LTX-2            & 60 & 13  & 98.21 & 95.78 & 99.01 & 62.86 & 69.31 & 98.83 & 98.42 & 88.92 \\
Open-Sora2.0     & 50 & 11  & 97.35 & 94.23 & 98.46 & 63.03 & 68.46 & 99.48 & 98.76 & 88.54 \\
Wan2.1-v1.1-Fun  & 20 & 1.5 & 94.12 & 94.78 & 99.02 & 66.72 & 69.55 & 98.18 & 99.31 & 88.81 \\
Ours-Prune       & 20 & 1.2 & 93.92 & 94.55 & 98.96 & 65.25 & 68.35 & 98.10 & 99.22 & 88.34 \\

\midrule
\multicolumn{11}{c}{\textbf{Bullet Time}} \\
\midrule
Wan2.1   & 20 & 14  & 95.33 & 95.79 & 99.09 & 66.65 & 70.56 & 98.16 & 99.30 & 89.27 \\
Ours-SFT & 20 & 1.2 & 94.01 & 94.86 & 99.00 & 66.85 & 69.29 & 98.15 & 99.34 & 88.79 \\
\rowcolor{gray!12}
Ours     & 4  & 1.2 & 93.46 & 94.06 & 98.76 & 64.76 & 70.71 & 98.15 & 98.58 & 88.35 \\

\midrule
\multicolumn{11}{c}{\textbf{Dolly Zoom}} \\
\midrule
Wan2.1   & 20 & 14  & 98.56 & 97.82 & 99.36 & 65.45 & 70.45 & 98.54 & 99.52 & 89.96 \\
Ours-SFT & 20 & 1.2 & 98.22 & 96.18 & 99.11 & 63.90 & 69.22 & 98.21 & 99.12 & 89.14 \\
\rowcolor{gray!12}
Ours     & 4  & 1.2 & 98.19 & 95.96 & 99.28 & 64.18 & 69.84 & 98.30 & 99.33 & 89.30 \\

\midrule
\multicolumn{11}{c}{\textbf{Slow Motion}} \\
\midrule
Wan2.1   & 20 & 14  & 96.48 & 96.01 & 99.20 & 63.42 & 68.15 & 97.36 & 98.97 & 88.51 \\
Ours-SFT & 20 & 1.2 & 96.65 & 95.57 & 98.95 & 62.29 & 67.89 & 97.55 & 99.25 & 88.31 \\
\rowcolor{gray!12}
Ours     & 4  & 1.2 & 95.82 & 94.98 & 99.03 & 61.52 & 68.75 & 97.28 & 99.00 & 88.05 \\

\bottomrule
\end{tabular}
}
\end{table*}

\paragraph{Baselines \& Metrics}

We conduct a systematic evaluation of \abbr in terms of video generation performance and efficiency. 
For generation performance, we combine VBench~\citep{Huang_2024_CVPR} and human evaluation to assess the generated videos from complementary perspectives.
Specifically, we use VBench to evaluate general video quality and compare \abbr with representative open-source video generation models, including HunyuanVideo-1.5~\citep{wu2025hunyuanvideo15technicalreport}, LTX2.3~\citep{hacohen2026ltx2efficientjointaudiovisual}, and Open-Sora2.0~\citep{zheng2026opensora20trainingcommerciallevel}. 
Since cinematic motion effects require effect-specific motion patterns that are not fully captured by standard automated metrics, we further conduct a human evaluation.
Specifically, we collect 50 video samples of bullet time and ask 10 participants to evaluate them. All videos are anonymized, randomly shuffled, and presented without revealing the identity of the corresponding generation model. In this evaluation, \abbr is compared with strong commercial video generation systems, including Kling, Hailuo, and Vidu, based on human preference scores.
For efficiency, we analyze the reductions in denoising time and FLOPs achieved by \abbr across different acceleration stages.

\paragraph{Deployment Platform}
We evaluate on-device deployment on an Infinix NOTE 60 Ultra equipped with the MediaTek Dimensity 8400 Ultimate 5G platform~\cite{infinix_note60ultra}. 
This SoC integrates an all-big-core Arm Cortex-A725 CPU, an Arm Mali-G720 MC7/MP7 GPU, and a MediaTek NPU 880~\cite{mediatek_dimensity8400}. 
The Mali-G720 MP7 GPU has a reported theoretical FP32 throughput of 2329.6 GFLOPS, approximately 2.33 TFLOPS~\cite{nanoreview_dimensity8400}. 
For reference, this theoretical FP32 peak is close to that of an NVIDIA GeForce GTX 960, whose FP32 throughput can be estimated as approximately 2.41 TFLOPS from its official CUDA core count and boost clock~\cite{nvidia_gtx960}. 
All mobile latency and memory measurements are conducted on this device unless otherwise specified.

\begin{figure*}[!t]
    \centering
    \includegraphics[width=\textwidth]{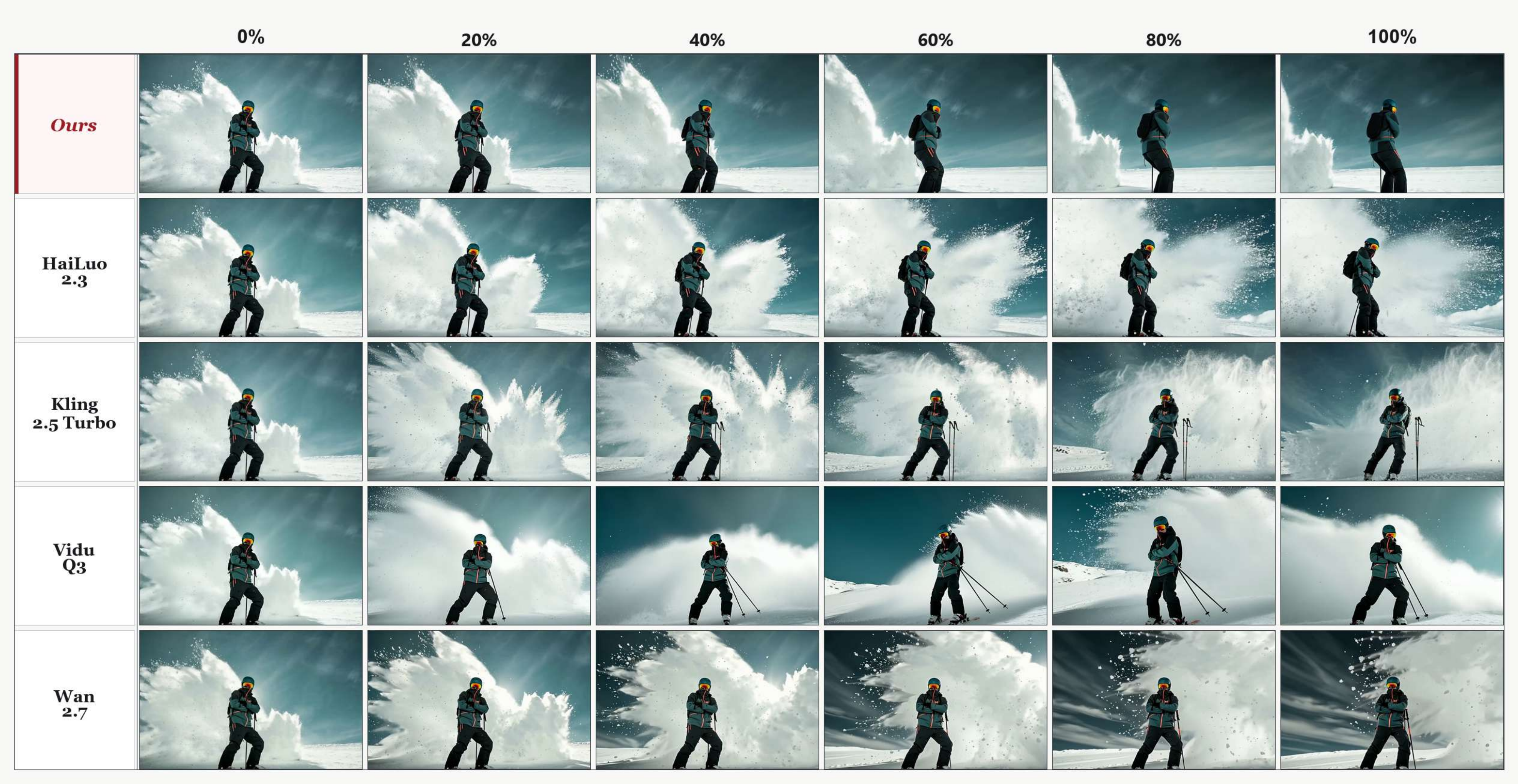}
    \caption{Qualitative comparison with commercial video generation models on cinematic motion effects. Frames are uniformly sampled at different temporal ratios from generated videos. Our method shows more stable camera progression and stronger subject-background consistency.}
    \label{fig:ratio_comparison_q3}
\end{figure*}

\subsection{Main Results}

\paragraph{Quantitative analysis.} 
We evaluate CineMobile with VBench~\citep{Huang_2024_CVPR} under three cinematic scenarios. 
As shown in \Cref{tab:vbench}, CineMobile achieves video quality comparable to the teacher model and other larger models, despite using only 4 denoising steps and a compact model size. 
Compared with the 20-step Wan2.1-I2V-14B teacher model, CineMobile uses only 10\% of the parameters and 20\% of the denoising steps, while limiting the absolute total-score gap to 0.92 on bullet time, 0.66 on dolly zoom, and 0.46 on slow motion. 
A closer look at individual metrics shows that CineMobile preserves motion smoothness and I2V consistency particularly well, suggesting that the 4-step student maintains reliable temporal dynamics and image-condition alignment. 
Overall, these results demonstrate that CineMobile substantially reduces inference cost while retaining competitive cinematic video generation quality.

\paragraph{Qualitative analysis.}
\Cref{fig:effects} and 
\Cref{fig:ratio_comparison_q3} presents a qualitative comparison using uniformly sampled frames across the generated video. Our method produces smoother camera progression and more coherent temporal evolution throughout the sequence.
By contrast, several commercial models exhibit stronger temporal fluctuation, less stable subject scale, or less consistent background motion, especially at later temporal ratios. 

\begin{figure}[t]
    \centering
    \begin{minipage}[t]{0.48\textwidth}
        \centering
        \includegraphics[width=\linewidth]{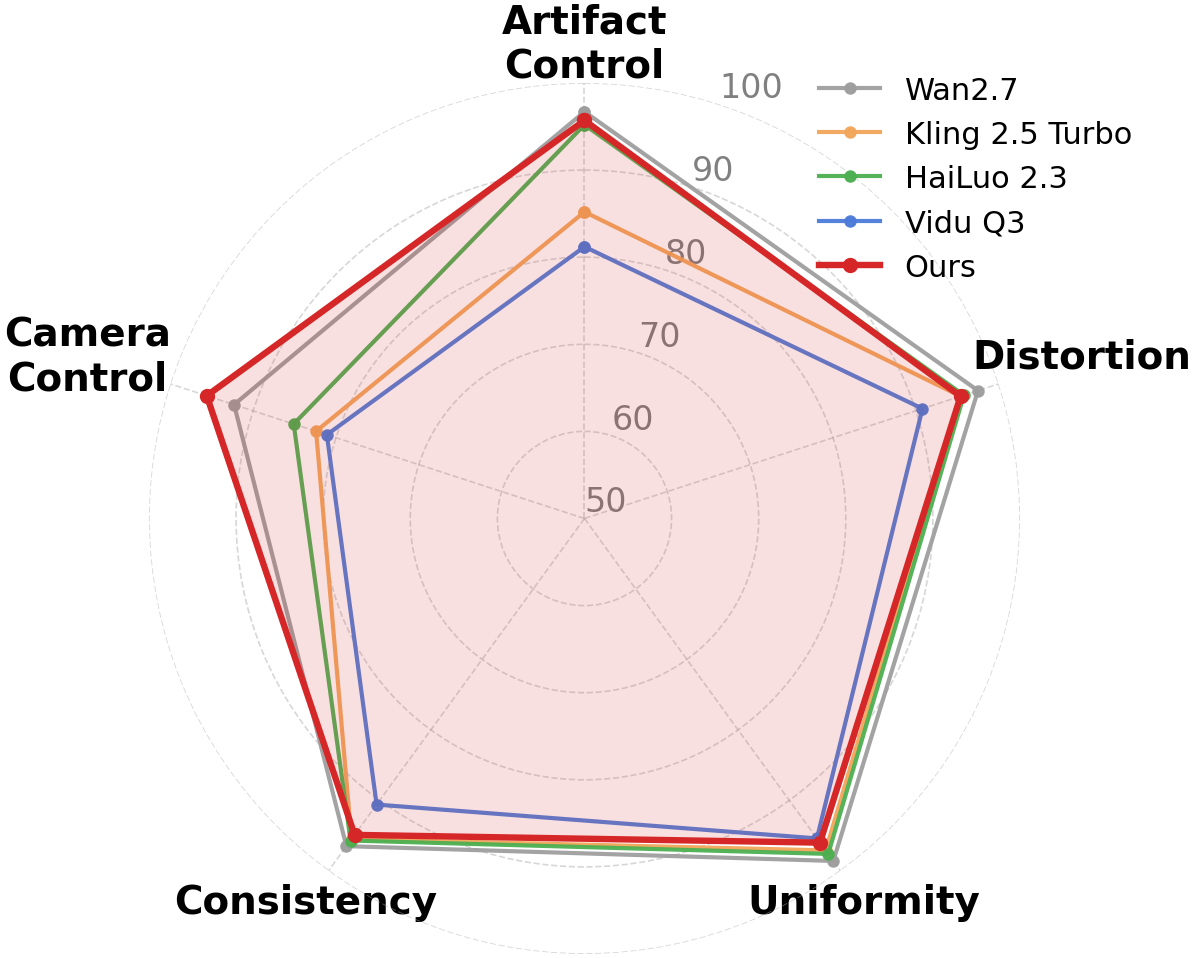}
        \captionof{figure}{
        Human evaluation of bullet time video generation against representative commercial models. 
        \abbr\ shows balanced quality with a clear advantage in camera control, while remaining competitive in overall quality.
        }
        \label{fig:commercial_models_radar}
    \end{minipage}
    \hfill
    \begin{minipage}[t]{0.48\textwidth}
        \centering
        \includegraphics[width=\linewidth]{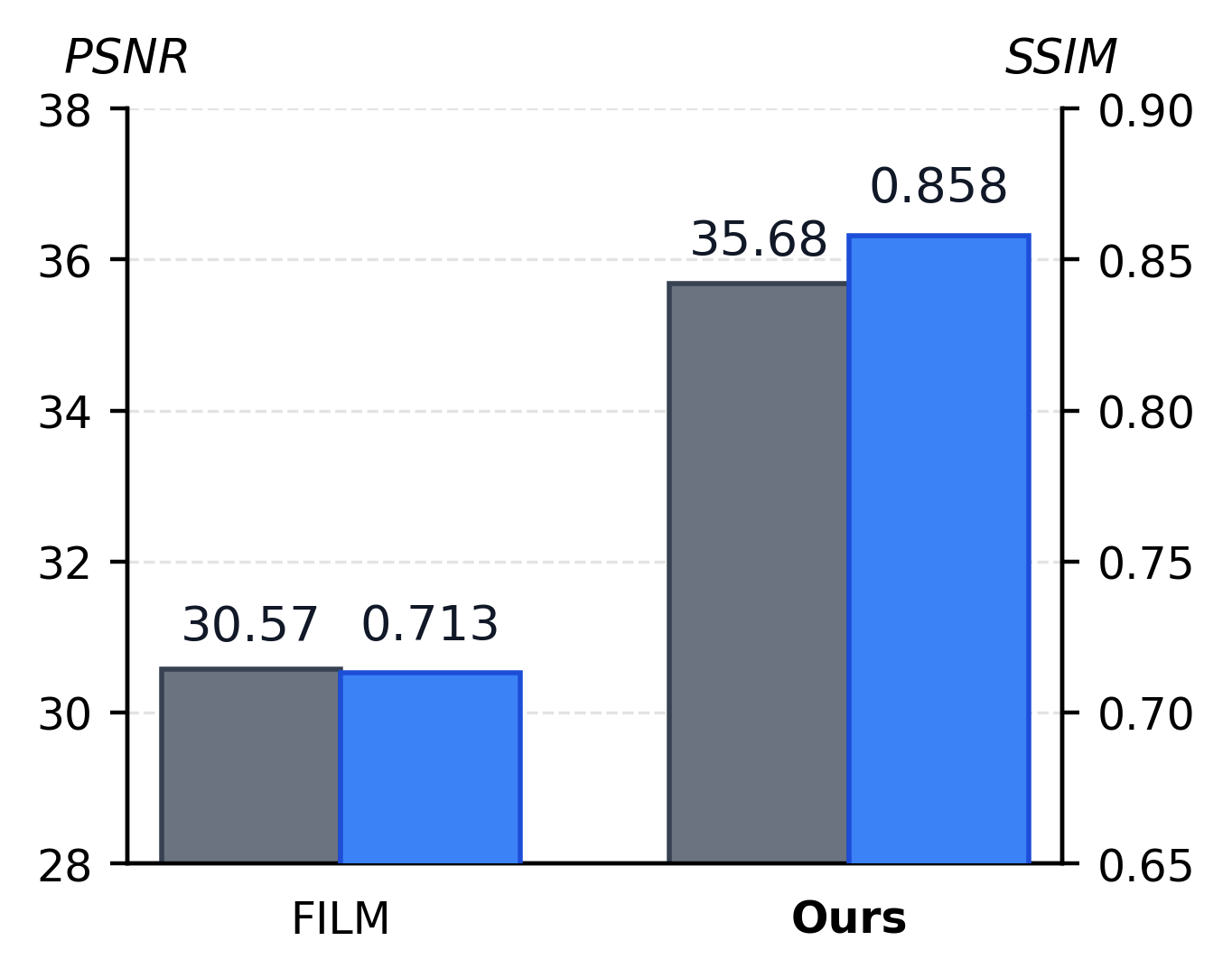}
        \captionof{figure}{
        Comparison with FILM~\citep{reda2022film} on $20\times$ slow motion synthesis. 
        Given the first and last frames, \abbr\ achieves consistently higher PSNR and SSIM than FILM, showing better reconstruction over large temporal gaps.
        }
        \label{fig:slow_motion_result}
    \end{minipage}
\end{figure}

We further compare \abbr\ with commercial models through human evaluation, as shown in \Cref{fig:commercial_models_radar}.
Overall, \abbr\ performs competitively across multiple perceptual dimensions, including artifact control, distortion, uniformity, and temporal consistency. It also achieves the strongest result in camera control, showing its advantage in controllable bullet-time motion. These results indicate that \abbr\ remains on par with representative commercial models in overall perceptual quality, while providing better camera controllability.

Overall, our model remains competitive across multiple perceptual dimensions, including artifact control, distortion, uniformity, and temporal consistency. Most notably, \abbr\ achieves the best score on camera control, which is consistent with our objective of optimizing cinematic motion effects rather than only static image quality. Although some commercial models obtain slightly stronger scores on individual dimensions, our method provides a favorable overall trade-off between controllable motion and perceptual quality. This is particularly notable given that \abbr\ is designed for efficient on-device deployment rather than cloud-scale generation.


\paragraph{Efficiency analysis.}
We further analyze the efficiency of \abbr\ across different acceleration stages in \Cref{fig:speed}. 
Structured pruning decreases the runtime to $12.28$ s by removing redundant DiT blocks. 
The most significant gain comes from step distillation, which reduces the denoising time to $2.45$ s and the DiT-FLOPs to $4.66\times10^{14}$, showing that reducing the number of denoising steps is the key factor for acceleration. 
Hybrid precision quantization brings a slight additional runtime improvement and mainly improves deployment efficiency by reducing memory cost. 
Overall, \abbr\ achieves $40.11\times$ faster DiT denoising and $71.89\times$ lower DiT-FLOPs than the teacher.

\subsection{Ablation \& Discussion}
\paragraph{Effect of GRPO refinement.}
We evaluate the effect of GRPO under the same 4-step sampling setting. 
As shown in Table~\ref{tab:grpo_ablation}, GRPO improves the total score for both the base model and \abbr, increasing the score by $+0.52$ and $+0.47$, respectively. 
The gains mainly come from perceptual-related metrics, including aesthetic quality and imaging quality, while other consistency metrics remain largely comparable. 
This trend suggests that GRPO helps refine the distilled generator by providing additional reward guidance beyond supervised distillation. 
Overall, GRPO improves the perceptual quality of 4-step generation without noticeably degrading image-conditioned consistency.

\begin{table*}[t]
\centering
\normalsize
\setlength{\tabcolsep}{3.2pt}
\renewcommand{\arraystretch}{1.12}
\caption{Ablation study of GRPO on bullet time generation. We compare models trained with and without GRPO on both the base model and CineMobile under the same sampling setting. GRPO benefits the distillation process and improves aesthetic quality and imaging quality.}
\label{tab:grpo_ablation}
\resizebox{\textwidth}{!}{
\begin{tabular}{lccccccccc}
\toprule
\textbf{Model} 
& \makecell{\textbf{Subject}\\\textbf{Consis.}} 
& \makecell{\textbf{Background}\\\textbf{Consis.}} 
& \makecell{\textbf{Motion}\\\textbf{Smooth.}} 
& \makecell{\textbf{Aesthetic}\\\textbf{Quality}} 
& \makecell{\textbf{Imaging}\\\textbf{Quality}} 
& \makecell{\textbf{I2V}\\\textbf{Subject}} 
& \makecell{\textbf{I2V}\\\textbf{Background}} 
& \makecell{\textbf{Total}\\\textbf{Score}} 
& \makecell{$\Delta$ \textbf{Total}} \\
\midrule

Base model (w/o GRPO) 
& 94.46 & 94.91 & 98.84 & 64.18 & 68.69 & 98.11 & 98.58 & 88.25 & -- \\

\textbf{Base model (w/ GRPO)} 
& \textbf{95.01} & 94.36 & \textbf{99.04} & \textbf{66.45} & \textbf{69.29} & \textbf{98.15} & \textbf{99.10} & \textbf{88.77} & \textbf{+0.52} \\

CineMobile (w/o GRPO) 
& 93.54 & 93.21 & \textbf{99.00} & 64.35 & 68.90 & 97.36 & \textbf{98.77} & 87.88 & -- \\

\textbf{CineMobile (w/ GRPO)} 
& 93.46 & \textbf{94.06} & 98.76 & \textbf{64.76} & \textbf{70.71} & \textbf{98.15} & 98.58 & \textbf{88.35} & \textbf{+0.47} \\

\bottomrule
\end{tabular}
}
\end{table*}

\paragraph{Comparison with frame interpolation.}
We compare \abbr\ with FILM~\citep{reda2022film} under a $20\times$ slow-motion setting, where only the first and last frames are provided as inputs. 
As shown in \Cref{fig:slow_motion_result}, \abbr\ achieves higher reconstruction quality, improving PSNR from 30.57 dB to 35.68 dB and SSIM from 0.713 to 0.858. 
This suggests that directly synthesizing slow-motion dynamics is more effective under large temporal gaps than conventional frame interpolation. 
We also observe that FILM can suffer from object interpenetration and motion inconsistency in this setting, while \abbr\ produces more coherent intermediate motion.

\section{Conclusion}

In this paper, we presented \abbr, an efficient on-device image-to-video generation framework for cinematic motion effects. 
Experiments on bullet time, dolly zoom, and slow motion demonstrate that \abbr substantially improves inference efficiency while maintaining competitive visual quality and temporal consistency relative to the teacher model, and showing favorable camera-motion control against commercial references in human evaluation. Overall, \abbr offers a new and practical solution for on-device deployment of image-to-video generation models. In future work, CineMobile can be further extended to a broader range of video generation models and cinematic effect generation tasks.

\setlength{\bibsep}{5pt}
\bibliography{reference}
\bibliographystyle{plainnat}

\newpage
\appendix

\section{Effect Definitions}

In this work, we focus on three representative cinematic motion effects: bullet time, dolly zoom, and slow motion. We briefly define them below and clarify the visual properties that are emphasized in our experiments.

\paragraph{Bullet Time.}
Bullet time refers to a cinematic effect in which the viewpoint appears to move rapidly around a relatively frozen subject or scene, producing a strong sense of spatial immersion and three-dimensional geometry. In video generation, this effect requires coherent camera motion, stable subject identity, and consistent scene structure across viewpoints. A successful bullet-time video should preserve the foreground subject while producing a smooth orbital or sweeping camera trajectory around it.

\paragraph{Dolly Zoom.}
The Hitchcock effect, also known as dolly zoom, is a classic camera effect created by simultaneously changing camera position and focal scale so that the subject remains relatively stable in size while the background perspective changes dramatically. In generated videos, this effect is characterized by a distinct push-pull perception: the foreground subject remains visually anchored, while the background expands or contracts in a way that induces strong spatial tension. This motion pattern places high demands on foreground-background consistency and controllable camera transformation.

\paragraph{Slow Motion.}
Slow motion refers to the visual effect of temporally stretching motion so that dynamic events appear to unfold more slowly than in real time. In our setting, this effect emphasizes smooth temporal interpolation, stable object appearance, and reduced motion artifacts across consecutive frames. Compared with bullet time and dolly zoom, slow motion is less dependent on explicit camera trajectory control and more sensitive to temporal continuity, detail preservation, and motion realism.

\section{Evaluation Criteria}

For automatic evaluation, we report the dimensions most relevant to cinematic motion effects: subject consistency, background consistency, motion smoothness, aesthetic quality, imaging quality, I2V subject, and I2V background. Together they cover the core requirements of our task, namely preserving subject identity, maintaining background coherence, and producing temporally smooth, visually plausible motion.

We omit several automatic metrics that are poorly aligned with our setting. The dynamic degree metric, for example, measures the overall magnitude of motion, whereas our target effects are not defined by motion amplitude. Bullet time and dolly zoom rely mainly on coordinated camera transformation and foreground--background geometric consistency, while slow motion emphasizes temporal smoothness and motion realism at reduced apparent speed. A higher dynamic degree therefore does not imply better effect quality, and can bias evaluation toward motion amplitude rather than controllable cinematic behavior.

We also exclude the VBench camera-control metric from the automatic comparison with base I2V models. There, we compare \abbr mainly with Wan2.1-based models that are not optimized for bullet time, dolly zoom, or slow motion. In this case, the camera-control score is hard to read as a fair measure of effect quality, since it largely reflects whether a model has been explicitly adapted to these motion patterns. We therefore rely on the selected VBench dimensions for automatic comparison, and assess effect-specific camera behavior through qualitative results and human evaluation.

For human evaluation against commercial models, we rate five dimensions. \emph{Artifact control} captures the absence of visible rendering artifacts, such as flicker, ghosting, unstable texture, or implausible structure. \emph{Distortion} measures whether the video preserves plausible scene geometry and avoids unnatural deformation during motion. \emph{Uniformity} measures appearance stability across frames, covering style, illumination, color tone, and overall presentation. \emph{Consistency} measures broader temporal coherence: stable subject identity, continuous motion, and preserved foreground--background relations over time. \emph{Camera control} measures whether the video follows the intended motion pattern, such as smooth orbital motion for bullet time, coordinated perspective change for dolly zoom, or temporally stretched but natural motion for slow motion. We report the average over these five dimensions as the total human-evaluation score.

\section{Compare with Width Pruning}
\label{appendix:pruning}

\begin{figure}
    \centering
    \includegraphics[width=\linewidth]{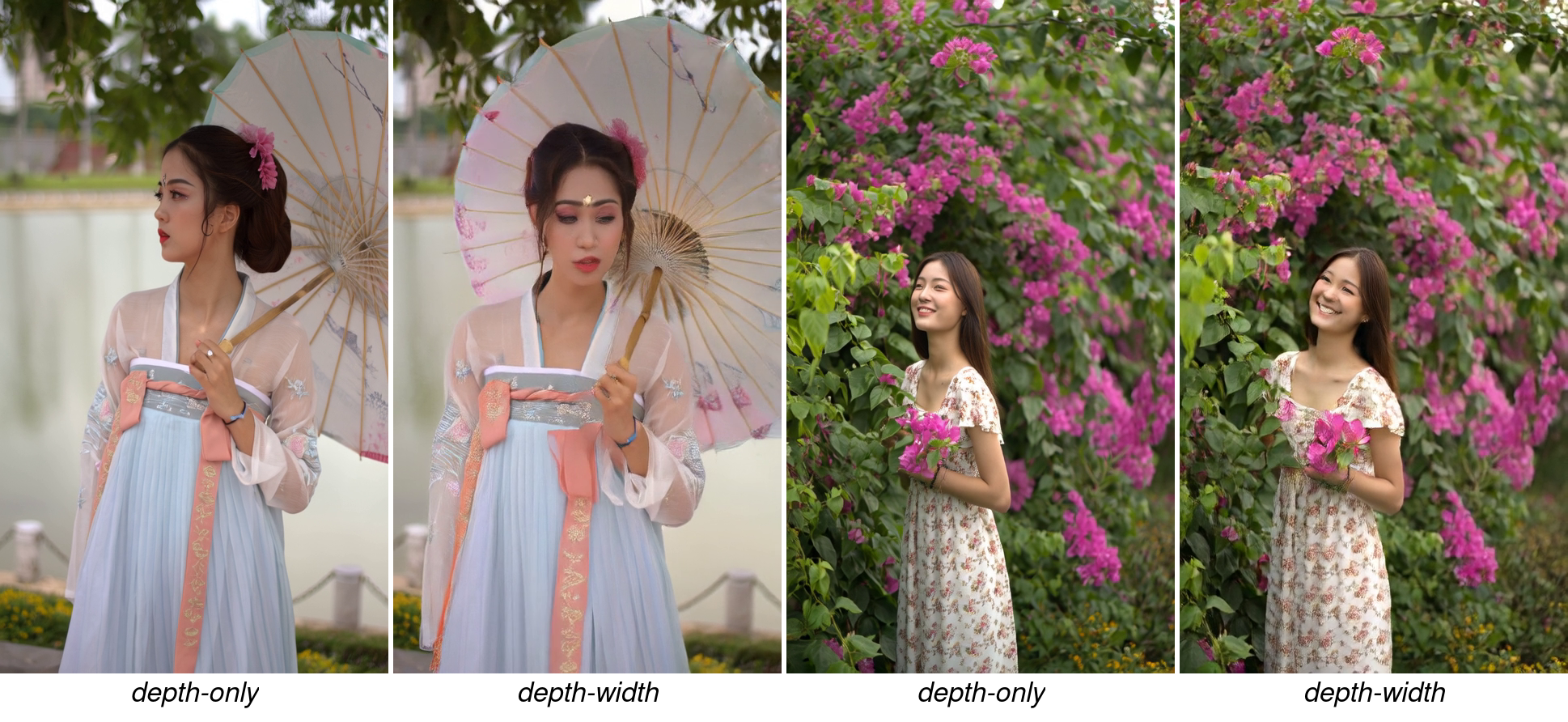}
    \caption{Width pruning may lead to identity shifts and deformation issues.}
    \label{fig:pruning}
\end{figure}

As illustrated in \Cref{fig:pruning}, the width-pruned method exhibits noticeable identity deviation and detail degradation. Therefore, instead of reducing the
hidden dimensionality, we keep the original backbone width unchanged and focus on structured depth pruning. This design confines compression to redundant layer
transformations while allowing the pruned model to remain aligned with the original backbone's feature space.

\begin{figure}[!t]
    \centering
    \includegraphics[width=\linewidth]{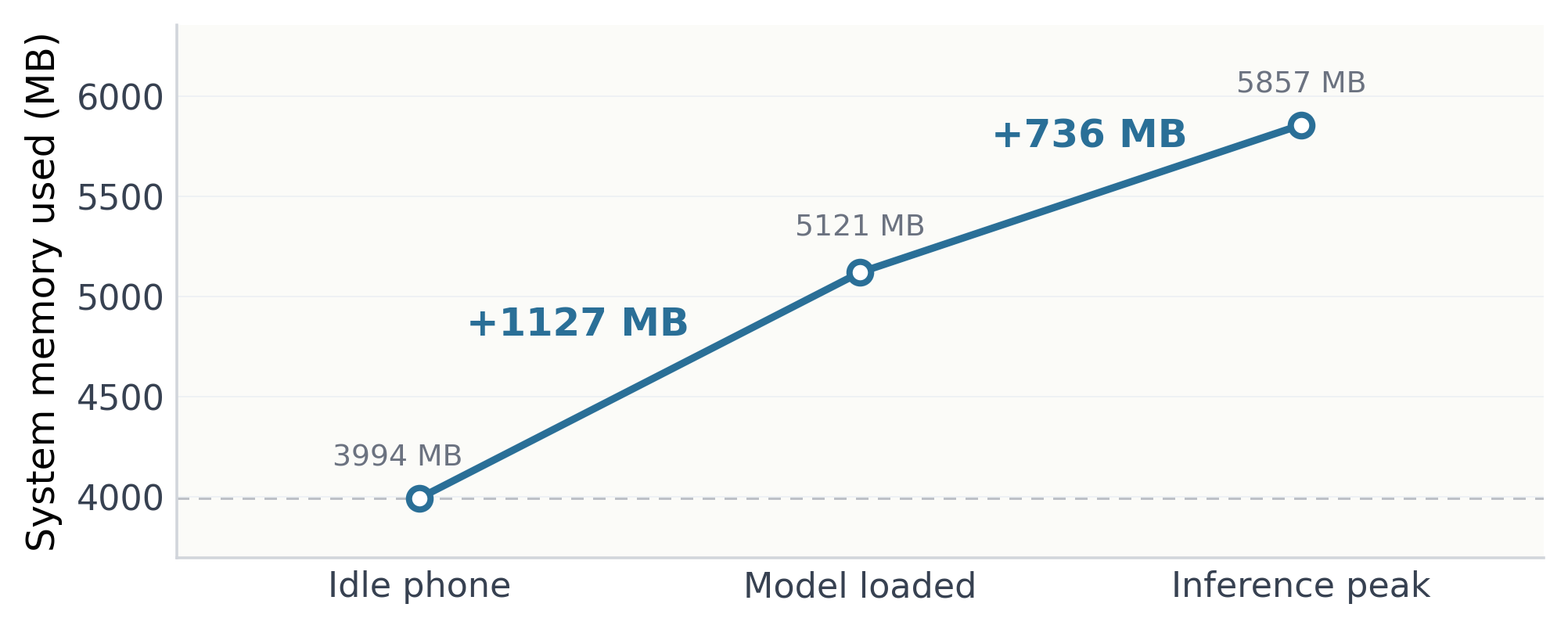}
    \caption{Memory footprint of CineMobile on an Infinix NOTE 60 Ultra. When the phone is idle under normal operation, the system memory usage is 3994 MB. After loading CineMobile, memory usage increases by 1127 MB to 5121 MB. During inference on 480p video, the peak memory usage reaches 5857 MB, requiring an additional 736 MB over the loaded-model state. Overall, CineMobile introduces only 1863 MB of additional runtime memory over the idle baseline, demonstrating that it can run efficiently on mobile devices with a modest memory footprint.}
    \label{fig:memory}
\end{figure}

\FloatBarrier
\clearpage
\section{Additional Qualitative Results}
\label{app:more_qualitative_results}

We provide additional qualitative results for the three cinematic motion effects considered in this work, namely bullet time, dolly zoom, and slow motion. All videos in this section are generated on a MediaTek Dimensity 8400 Ultimate 5G platform. For each example, frames are uniformly sampled at different temporal ratios from the generated video. These visualizations complement the main-paper comparisons by showing that \abbr\ maintains stable subject identity, coherent foreground-background relations, and smooth temporal progression across diverse portrait scenes.

\FloatBarrier
\section{Data Curation and Preprocessing}
Our training images are collected from high-quality portrait datasets, including PPR10K and Pixel. Since our task focuses on cinematic motion effects for portrait-centric image-to-video generation, we further apply a dedicated data curation pipeline to improve both visual quality and scene suitability. In particular, we use Qwen3-VL-30B-A3B-Instruct to automatically filter out images with overly simple or uninformative backgrounds, extreme overexposure or underexposure, insufficient spatial resolution, multiple prominent persons, overly tight face-dominant close-ups, missing human subjects, or very low aesthetic quality. We additionally remove samples with poor composition, ambiguous foreground-background structure, or limited motion potential, since such cases are less suitable for evaluating camera-driven effects such as bullet time and dolly zoom. This preprocessing procedure yields a cleaner and more task-aligned portrait dataset for both training and evaluation. All images used for both training and inference are sourced from publicly accessible datasets.

\begin{figure}[!t]
    \centering
    \includegraphics[width=\linewidth]{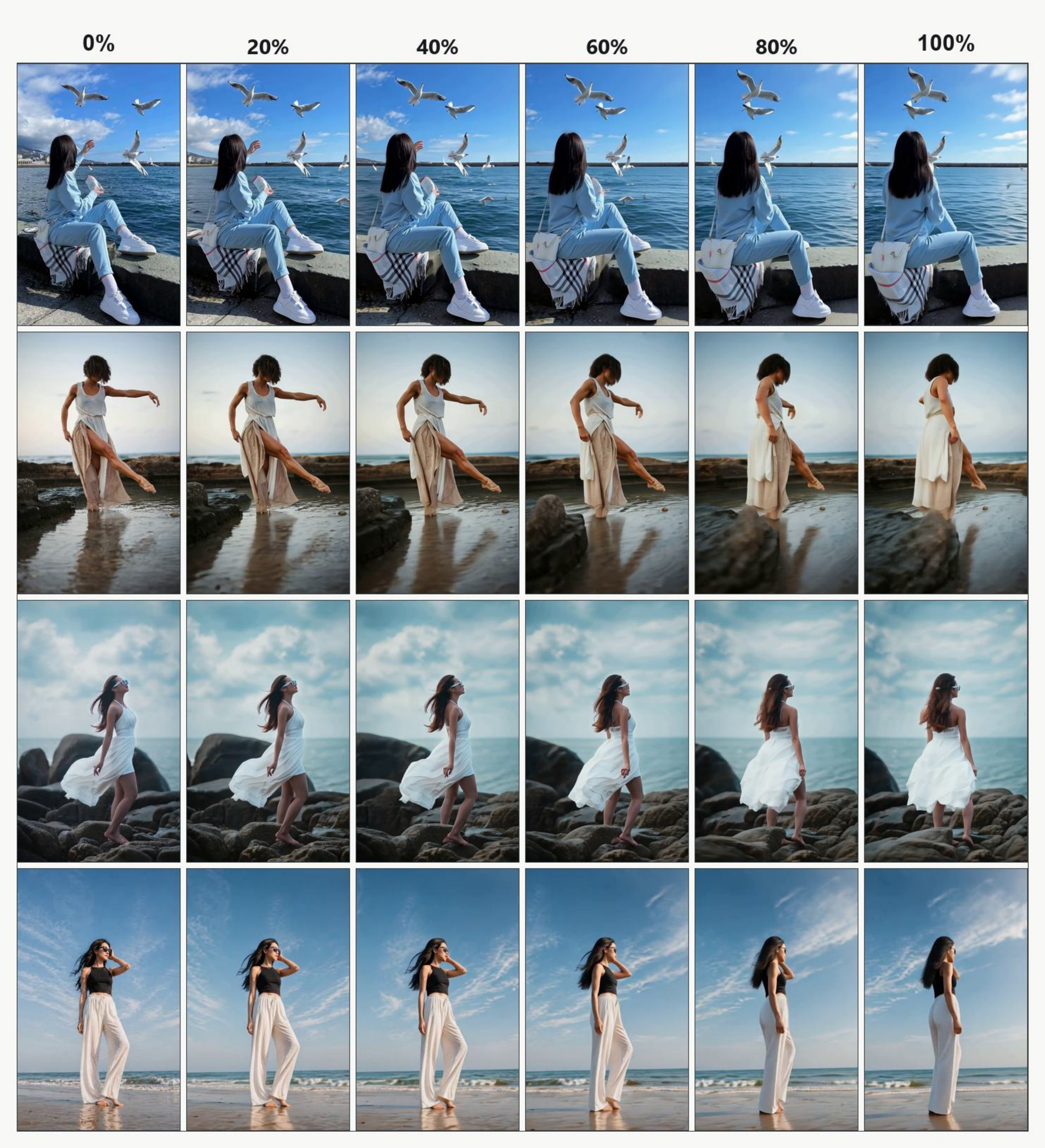}
    \caption{Additional qualitative results on \textbf{bullet time}. Each row shows one generated example, and columns correspond to uniformly sampled frames at different temporal ratios. \abbr\ produces smooth viewpoint transition while preserving subject appearance and scene geometry throughout the sequence.}
    \label{fig:appendix_bullet_time}
\end{figure}

\clearpage

\begin{figure}[t]
    \centering
    \includegraphics[width=\linewidth]{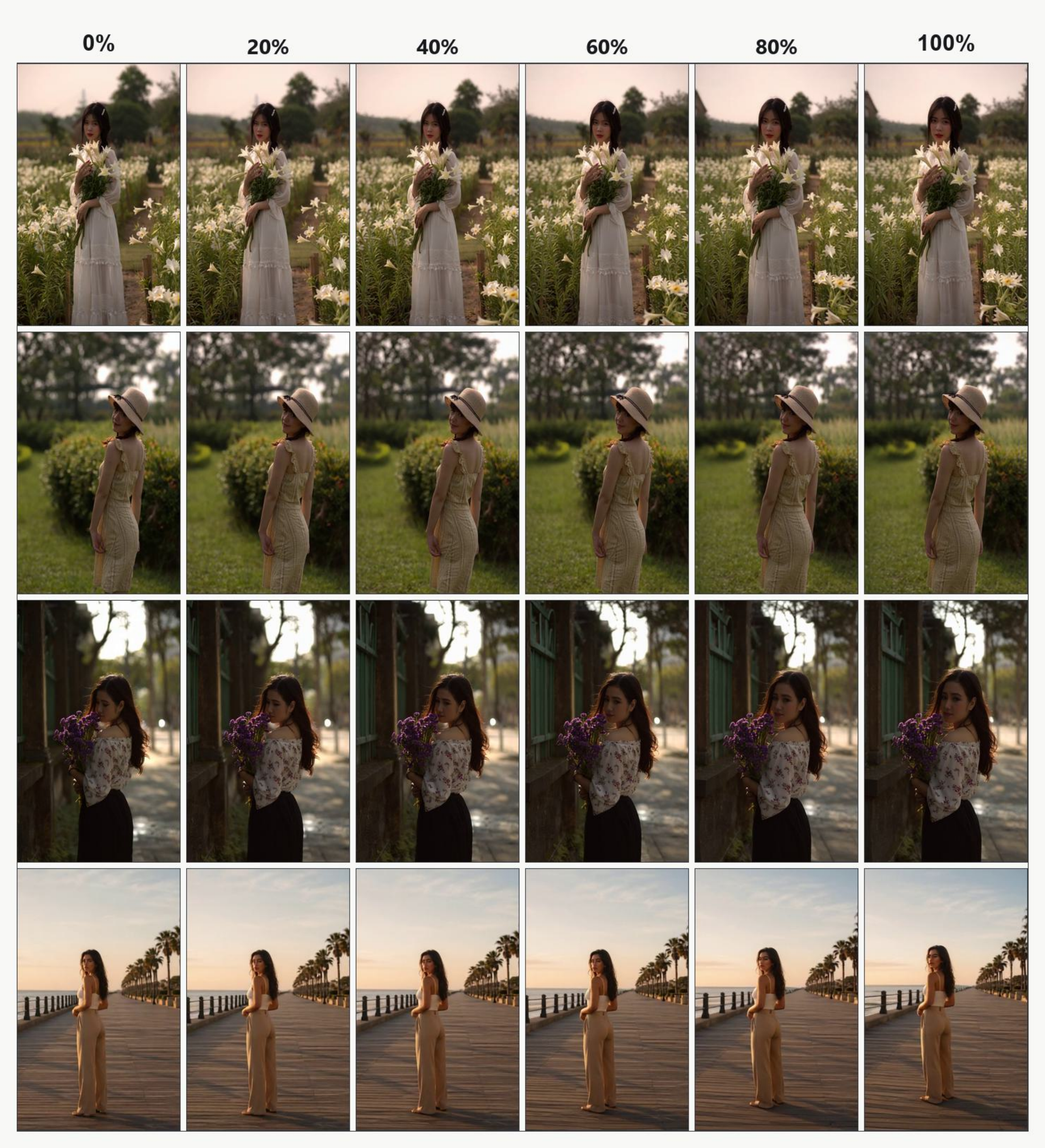}
    \caption{Additional qualitative results on \textbf{dolly zoom}. The sampled frames illustrate consistent foreground stabilization together with coordinated background perspective variation, which is the key visual characteristic of the Hitchcock dolly-zoom effect.}
    \label{fig:appendix_dolly_zoom}
\end{figure}

\clearpage

\begin{figure}[t]
    \centering
    \includegraphics[width=\linewidth]{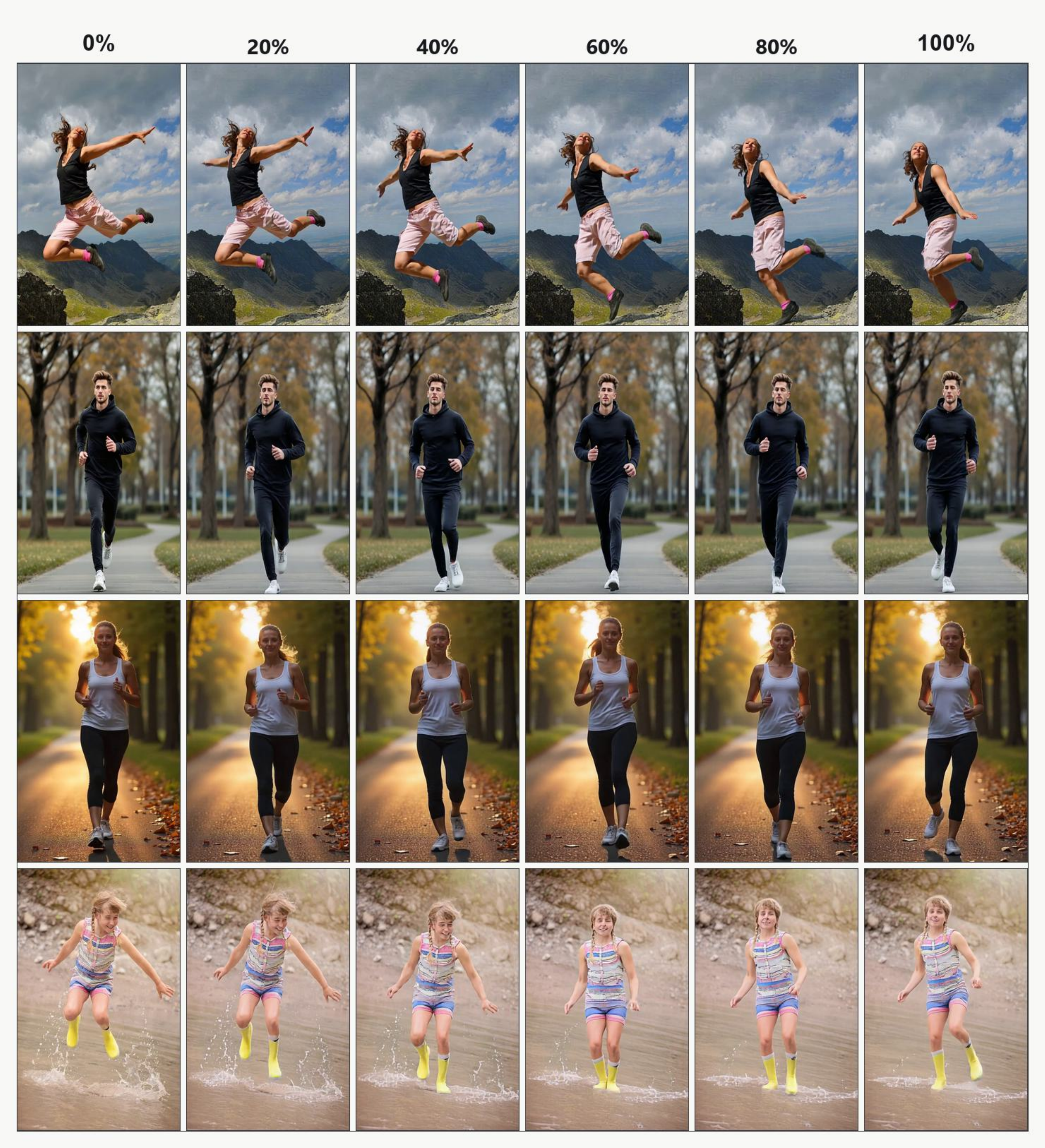}
    \caption{Additional qualitative results on \textbf{slow motion}. \abbr\ preserves temporal smoothness and subject consistency while reducing abrupt motion changes, yielding visually plausible slow-motion sequences across different scenes.}
    \label{fig:appendix_slow_motion}
\end{figure}

\newpage
\end{document}